\documentclass{article}

\usepackage[preprint,nonatbib]{nips_2018}

\usepackage[utf8]{inputenc} 
\usepackage[T1]{fontenc}    
\usepackage{hyperref}       
\usepackage{url}            
\usepackage{booktabs}       
\usepackage{amsfonts}       
\usepackage{nicefrac}       
\usepackage{microtype}      
\usepackage[numbers]{natbib}

\usepackage{graphicx}
\usepackage{subfig}
\usepackage{amsmath} 
\usepackage{epsfig}
\usepackage{breqn}
\usepackage{bm}
\usepackage{amssymb}
\usepackage{color}

\newcommand{\bb}{\bm}
\usepackage{algorithm,algorithmic}
\DeclareMathOperator*{\argminA}{arg\,min} 

\DeclareMathOperator*{\argmaxA}{arg\,max} 

\title{Learning Personalized Representation for Inverse Problems in Medical Imaging Using Deep Neural Network}

\author{
  Kuang Gong\\
Gordon Center for Medical Imaging\\
Massachusetts General Hospital\\
Department of Biomedical Engineering\\
 University of California, Davis\\
   \And
    Kyungsang Kim \\
Gordon Center for Medical Imaging\\
Massachusetts General Hospital\\
   \And
   Jianan Cui \\
Gordon Center for Medical Imaging\\
Massachusetts General Hospital\\
   \And
   Ning Guo \\
Gordon Center for Medical Imaging\\
Massachusetts General Hospital\\
   \And
   Ciprian Catana \\
Martinos Center for Biomedical Imaging\\
Massachusetts General Hospital\\
   \And
   Jinyi Qi \\
Department of Biomedical Engineering\\
 University of California, Davis\\
   \And
   Quanzheng Li\thanks{Correspondence to li.quanzheng@mgh.harvard.edu}  \\
Gordon Center for Medical Imaging\\
Massachusetts General Hospital\\
}

\begin{document}

\maketitle

\begin{abstract}
Recently deep neural networks have been widely and successfully applied in computer vision tasks and attracted growing interests in medical imaging. One barrier for the application of deep neural networks to medical imaging is the need of large amounts of prior training pairs, which is not always feasible in clinical practice. In this work we propose a personalized representation learning framework where no prior training pairs are needed, but only the patient's own prior images. The representation is expressed using a deep neural network with the patient's prior images as network input. We then applied this novel image representation to inverse problems in medical imaging in which the original inverse problem was formulated as a constraint optimization problem and solved using the alternating direction method of multipliers (ADMM) algorithm. Anatomically guided brain positron emission tomography (PET) image reconstruction and image denoising were employed as examples to demonstrate the effectiveness of the proposed framework. Quantification results based on simulation and real datasets show that the proposed personalized representation framework outperform other widely adopted methods.

\end{abstract}
\section{Introduction}
Over the past several years, deep neural networks have been widely and successfully applied to computer vision tasks such as image segmentation, object detection, and image super resolution, by demonstrating better performance than state-of-art methods when large amounts of datasets are available. For medical imaging tasks such as lesion detection and region-of-interest (ROI) quantification, obtaining high-quality diagnostic images is essential. 
Recently the neural network method has been applied to inverse problems in medical imaging\cite{wang2016accelerating,sun2016deep,kang2017deep,chen2017low,hammernik2018learning,wu2017iterative,adler2018learned,schlemper2018deep,gong2017iterative,kim2018penalized}.

During network training, images reconstructed from high dose or long-scanned duration are needed as training labels. However, collecting large amounts of training labels is not an easy task: high-dose computed tomography (CT) has potential safety concerns; long-scanned dynamic PET is not employed in routine clinical practice; in cardiac magnetic resonance imaging (MRI), it is impossible to acquire breath-hold and fully sampled 3D images. With limited amounts of high-quality patient datasets available, overfitting can be a potential pitfall: if a new patient dataset does not lie in the training space due to population difference, the trained network cannot accurately recover unseen structures. In addition, low-quality images are often simulated by artificially downsampling the full-dose/high-count data, which may not reflect the real physical conditions of low-dose imaging. This mismatch between training and the real clinical environment can reduce the network performance.

Apart from using training pairs to perform supervised learning, a lot of prior arts focus on exploiting prior images acquired from the same patient to improve the image quality. The priors can come from temporal information \citep{chen2008prior,guobao15}, different physics settings \citep{gnahm2015anatomically}, or even other imaging modalities \citep{gong2017direct}. They are included into the maximum posterior estimation or sparse representation framework using pre-defined analytical expressions or  pre-learning steps. The pre-defined expressions might not be able to extract all the useful information, and the pre-learnt model might not be optimal for the later inverse-problem optimization as no data-consistency constraint is enforced during pre-learning. Ideally the learning process should be included inside the inverse-problem solving process.

Recently, the deep image prior framework proposed in \citep{ulyanov2017deep} shows that convolutional neural networks (CNNs) can learn intrinsic structure information from the corrupted images based on random noise input. No prior training pairs or learning process are needed in this framework. The ability of neural networks to learn the structure information is also revealed in  generative adversarial networks (GANs) \citep{goodfellow2014generative,radford2015unsupervised}, where various distributions can be generated based on random noise input. These prior arts show that the network itself can be treated as a regularizer, and it is possible to train a network without high-quality reference images. Furthermore, it has been shown in conditional GAN works \citep{mirza2014conditional,isola2017image,zhu2017unpaired} that when the input is not random noise, but the associated prior information, the prediction results can be improved.

Inspired by the prior arts, we propose a new framework to solve inverse problems in medical imaging based on learning personalized representation. The personalized representation is learnt through a deep neural network with prior images of the same patient as network input. ADMM algorithm was employed to decouple the inverse-problem optimization and network training steps. Modified 3D U-net \citep{cciccek20163d} was employed as the network structure. The total number of trainable parameters for the network is about 1.46 million, which is less than the unknowns in the original inverse problem. 

PET is a molecular imaging modality widely used in neurology studies \cite{gong2016designing}. The image resolution of current PET scanners is still limited by various physical degradation factors \cite{gong2017sinogram}. Improving PET image resolution is essential for a lot of applications, such as dopamine neurotransmitter imaging, brain tumor staging and early diagnosis of Alzheimer's disease. For the past decades, various efforts are focusing on using MR or CT to improve PET image quality \cite{bai2013magnetic}. In this work, anatomically guided brain PET image reconstruction and denoising are employed as two examples to demonstrate the effectiveness of the proposed framework.

 The main contributions of this work include:(1) proposing to employ the patient's prior images as network input to construct a personalized representation;(2) embedding this personalized representation in complicated inverse problems to accurately estimate unknown images; (3) proposing a ADMM framework to decouple the whole optimization into a penalized inverse problem and a network training problem; (4) demonstrating the effectiveness of the proposed framework using clinical brain imaging applications.
 
\section{Methodology}
\subsection{Related works}
In inverse problems such as image reconstruction and denoising, the measured data $\bb{y} \in \mathbb{R}^{M \times 1} $ can be modelled as a collection of independent random variables and its mean $\bar{\bb{y}} \in \mathbb{R}^{M \times 1} $ is related to the original image $\bb{x} \in \mathbb{R}^{N \times 1}$ through an affine transform
\begin{equation}
\bar{\bb{y}} = \bb{A}\bb{x} + \bb{s},
\label{eqn_mean}
\end{equation}
where $\bb{A} \in \mathbb{R}^{M \times N}$ is the transformation matrix and $\bb{s} \in \mathbb{R}^{M \times 1}$ is the known additive item. For the past decades, a lot of efforts focus on representing the original image $\bb{x}$ by different basis functions \citep{do2005contourlet,elad2006image,guobao15}. For example, in dictionary learning proposed by \citet{elad2006image} , $\bb{x}_i \in \mathbb{R}^{P \times 1}$, the $i^{\text{th}}$ patch of $\bb{x}$, can be represented as
\begin{equation}
\bb{x}_i = \bb{D}\bb{\alpha}_i, 
\label{dl}
\end{equation}
where $\bb{D}\in \mathbb{R}^{P \times L}$ is the pre-learnt overcomplete dictionary based on high-quality reference images, and $\bb{\alpha}_i \in \mathbb{R}^{L \times 1}$ is the coefficient for the $i^{\text{th}}$ patch. $L_{0}$ constraint is added on $\bb{\alpha}_i$ to enforce sparse representation. In the kernel method proposed by \citet{guobao15}, the original image  $\bb{x}$ is represented as
\begin{equation}
\bb{x}= \bb{K}\bb{\alpha}, 
\label{kernel}
\end{equation}
where $\bb{K}  \in \mathbb{R}^{N \times N}$ is the kernel matrix calculated using the radial basis function, and $\bb{\alpha}\in \mathbb{R}^{N \times 1}$ is the kernel coefficient. Both the dictionary learning and the kernel method assumes the local/global of the original unknown image can be linearly represented using pre-learnt/pre-defined basis vectors.
In a recent work by \citet{gong2017iterative} for image reconstruction tasks, the original unknown image $\bb{x}$ is represented using CNN as
\begin{equation}
\bb{x}= f(\bb{\theta}|\bb{\alpha}), 
\label{eq:sparse-represent}
\end{equation}
where $f:\mathbb{R}\rightarrow\mathbb{R}$ represents the neural network, $\bb{\theta}$ are the parameters of the neural network, and $\bb{\alpha}$ denotes the network input. In this CNN representation, the network is pre-learnt using large number of training pairs from different patient datasets, network parameters $\bb{\theta}$ are thus fixed, and the network input $\bb{\alpha}$ is keeping updated during the reconstruction process. Recently \citet{ulyanov2017deep} uses the same representation as shown in (\ref{eq:sparse-represent}) for image super-resolution and denoising tasks. Different from \citeauthor{gong2017iterative}'s approach, the network input $\bb{\alpha}$ is a pre-defined random noise, and the network parameters $\bb{\theta}$ are keeping updated in the image restoration process. One biggest advantage of \citeauthor{ulyanov2017deep}'s framework is that pre-training steps and large number of training pairs are not needed.
\subsection{Personalized representation using deep neural network}
\begin{figure}[t]
\centering
\subfloat{\includegraphics[trim=1.8cm 0.2cm 1.8cm 0.3cm, clip, width=4in]{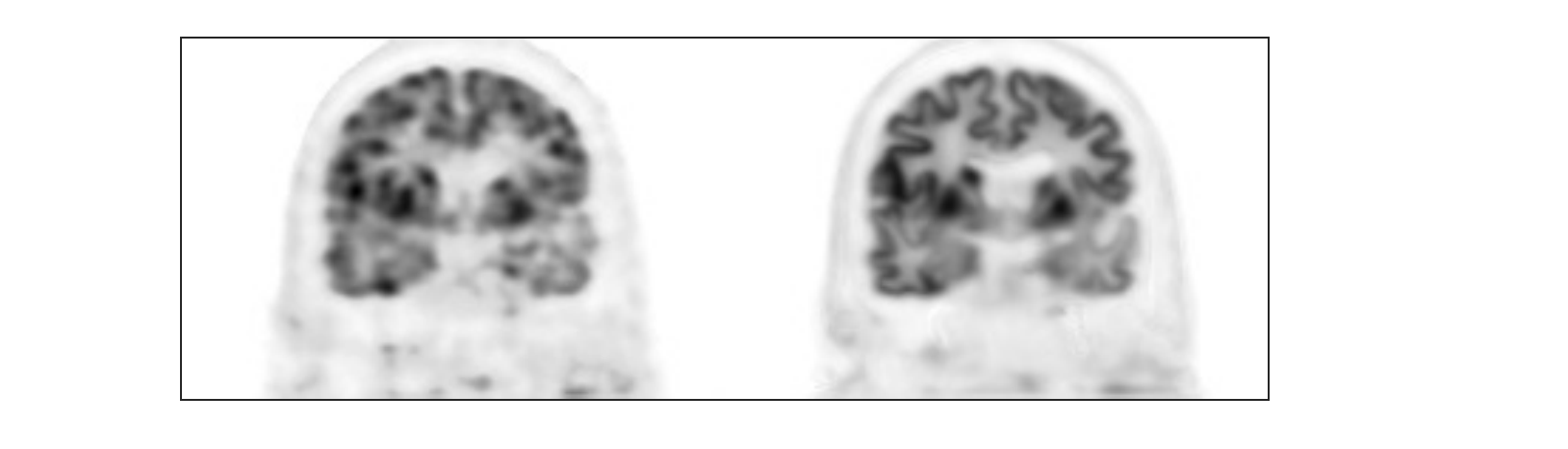}}
\caption{{\small{One coronal view of the network output by using random noise as network input (left) and using the Patient's MR image as network input (right). Details of the brain PET-MR data set are introduced in Section \ref{section:brain_pet}} } }
\label{fig:effect_of_conditional}
\end{figure}

Inspired by the prior arts, here we propose a personalized representation framework for inverse problems in medical imaging. In this proposed framework, the original unknown image $\bb{x}$ is represented by a deep neural network as shown in (\ref{eq:sparse-represent}). The prior image of the same patient is employed as network input $\bb{\alpha}$ to construct a personalized representation. The network parameters $\bb{\theta}$ are updated in the inverse-problem optimization process. Compared to employing random noise as network input as presented in \citep{ulyanov2017deep}, using prior images of the patient can make the representation more accurate. To illustrate this, Fig.~\ref{fig:effect_of_conditional} presents the comparison of network outputs with different network inputs using the brain dataset introduced in Section \ref{section:brain_pet}. Clearly, when the patient' own MR image is employed as network input, the cortices regions are clearer and noise in white matter regions is reduced. After substituting $\bb{x}$ with the neural network representation (\ref{eq:sparse-represent}), the original data model shown in (\ref{eqn_mean}) can be rewritten as  
\begin{equation}
\bar{\bb{y}} = \bb{A}f(\bb{\theta} | \bb{\alpha}) + \bb{s}.
\label{eqn_mean_alpha}
\end{equation}
 Supposing the measured random variable ${y}_i$ follows a distribution of $p(y_i|\bb{x})$, the log likelihood for the measured data $\bb{y}$ can be written as
\begin{equation}
L(\bb{y}|\bb{x}) = \log \sum\nolimits_{i=1}^M p(y_i | \bb{x}).
\label{likelihood}
\end{equation}
The maximum likelihood estimate of the unknown image $\bb{x}$ can be calculated in two steps as
\begin{align}
\hat{\bb{\theta}} = \mbox{arg} \max L(\bb{y}|\bb{\theta}), \hat{\bb{x}} = f(\hat{\bb{\theta}} | \bb{\alpha}).  \label{eq:max_alpha}
\end{align}

\subsubsection{Optimization}
The objective function in (\ref{eq:max_alpha}) is difficult to solve due to the coupling between the likelihood function and the neural network. Here we transfer it to the constrained format as below
\begin{equation}
\begin{aligned}
& {\text{max}}
& & L(\bb{y}|{\bb{x}})  \\
& \text{s.t.}
& & {\bb{x}} = f(\bb{\theta} | \bb{\alpha}). 
\end{aligned}
\label{eq:contrained_optim}
\end{equation}
We use the augmented Lagrangian format for the constrained optimization 
problem in (\ref{eq:contrained_optim}) as
\begin{equation}
L_{\rho} = L(\bb{y}|{\bb{x}}) - \frac{\rho}{2} \Vert{\bb{x}} -{f}({\bb{\theta}} | \bb{\alpha})+ \bb{\mu}\Vert^2 + \frac{\rho}{2} \Vert \bb{\mu} \Vert^2,
\label{admm_scale}
\end{equation}
which can be solved by the ADMM algorithm iteratively in three steps
\begin{align}
\bb{\theta}^{n+1} &= \argminA_{\bb{\theta}} \Vert {f}({\bb{\theta}} | \bb{\alpha})- (\bb{x}^{n} + \bb{\mu}^n)\Vert^2,  \label{eq:sub1} \\
\bb{x}^{n+1} &= \argmaxA_{\bb{x}}  L(\bb{y}|\bb{x}) - \frac{\rho}{2} \Vert\bb{x} - {f}({\bb{\theta}^{n+1}} | \bb{\alpha}) + \bb{\mu}^n\Vert^2, \label{eq:sub2}\\
\bb{\mu}^{n+1} &= \bb{\mu}^n + \bb{x}^{n+1} - f({\bb{\theta}^{n+1}} | \bb{\alpha}).  \label{eq:sub3}
\end{align} 
After using the ADMM algorithm, we decoupled the constraint optimization problem into a network training problem with L2 norm as loss function (\ref{eq:sub1}), and a penalized inverse problem (\ref{eq:sub2}). For subproblem (\ref{eq:sub1}), currently network training is mostly based on first order methods, such as the Adam algorithm \citep{kingma2014adam} and the Nesterov's accelerated gradient (NAG) algorithm \citep{nesterov1983method}. The L-BFGS algorithm \citep{liu1989limited} is a quasi-newton method, combining a history of updates to approximate the Hessian matrix. It is not widely used in network training as it requires large batch size to accurately calculate the descent direction, which is less effective than first order methods for large-scale applications. In this proposed framework, as only the patient's own prior images are employed as network input, the data size is much smaller than the traditional network training. In our case, the L-BFGS method is preferred to solve subproblem (\ref{eq:sub1}) due to its stability and better performance observed in the experiments (results shown in Section.~\ref{section:implement_details}). For the penalized inverse problem shown in subproblem (\ref{eq:sub2}), it is application dependent and is discussed as below.
\paragraph{PET image reconstruction}
In PET image reconstruction, $\bb{A}$ is the detection probability matrix, with $A_{ij}$ denoting the probability of photons originating from voxel $j$ being detected by detector $i$. $\bb{s} \in \mathbb{R}^{M \times 1}$ denotes the expectation of scattered and random events. $M$ is the number of lines of response (LOR). The measured photon coincidences follow Poisson distribution, and the log-likelihood function $L(\bb{y}|\bb{x})$ can be explicitly written as
\begin{equation}
L(\bb{y}|\bb{x}) = \log \sum\nolimits_i p(y_i | \bb{x}) =\sum\nolimits_{i=1}^M y_i \log \bar{y}_i - \bar{y}_i - \log y_i!.
\label{likelihood_explicitly}
\end{equation}
Subproblem (\ref{eq:sub2}) thus corresponds to a penalized reconstruction problem. The optimization transfer method \cite{lange2000optimization} is used to solve it. As $\bb{x}$ in $L(\bb{y}|\bb{x})$ is coupled together, we first construct a surrogate function $Q_{L}(\bb{x}|\bb{x}^n)$ for $L(\bb{y}|\bb{x})$ to decouple the image pixels so that each pixel can be optimized independently. $Q_{L}(\bb{x}|\bb{x}^n)$ is constructed as follows
\begin{equation}
Q_{L}(\bb{x}|\bb{x}^n) =  \sum\nolimits_{j=1}^{n_j}A_{.j}(\hat{x}^{n+1}_{j,\text{EM}}\log x_j - x_j),
\label{admm_scale}
\end{equation}
where $A_{.j} = \sum\nolimits_{i=1}^{M}A_{ij}$ and $\hat{x}^{n+1}_{j,\text{EM}}$ is calculated by 
\begin{equation}
\hat{x}^{n+1}_{j,\text{EM}} = \frac{x^n_j}{A_{.j}} \sum\nolimits_{i=1}^{M}A_{ij}\frac{y_i}{ [\bb{A}\bb{x}^n]_i + s_i}.
\label{admm_scale}
\end{equation}
It can be verified that the constructed surrogate function $Q_{L}(\bb{x}|\bb{x}^n)$ fulfills the following two conditions:
\begin{align}
Q_{L}(\bb{x}; \bb{x}^n)-Q_{L}(\bb{x}^n; \bb{x}^n) &\leq  L(\bb{y};\bb{x})-L(\bb{y}; \bb{x}^n), \\
\nabla Q_{L}(\bb{x}^n; \bb{x}^n) &= \nabla L(\bb{y}; \bb{x}^n).
\label{opt_transfer_cond}
\end{align}
After getting this surrogate function, subproblem (\ref{eq:sub1}) can be optimized pixel by pixel. For pixel $j$, the surrogate objective function for subproblem (\ref{eq:sub1}) is 
\begin{align}
P(x_j|\bb{x}^n)  &=  A_{.j}(\hat{x}^{n+1}_{j,\text{EM}}\log x_j - x_j)  - \frac{\rho}{2}\left[x_j - f(\bb{\theta}|{\bb{\alpha}})^n_j + {\mu}_j^n\right]^2. 
\label{eq:surrogate-pixel}
\end{align}
The final update equation for pixel $j$ after maximizing (\ref{eq:surrogate-pixel}) is 
\begin{align}
{x}^{n+1}_{j}  &= \frac{1}{2}[f(\bb{\theta}^n| \bb{\alpha})_j - \mu^n_j -A_{.j} / \rho] + \frac{1}{2}\sqrt{[f(\bb{\theta}^n | \bb{\alpha})_j - \mu^n_j - A_{.j}/ \rho]^2 + 4\hat{x}^{n+1}_{j,\text{EM}} A_{.j} /\rho}.
\label{update_sub1}
\end{align}
\paragraph{PET image restoration}
For PET image denoising problems, $\bb{A}$ is an identity matrix and the noisy images follow the Gaussian distribution. Thus objective function (\ref{eq:max_alpha}) can be seen as a network training problem, which can be directly solved using the L-BFGS algorithm with L2 norm as the loss function. For PET image deblurring problems, $\bb{A}$ is a blurring matrix. Solving the optimization problem follows the same flowchart as shown in the previous mentioned PET image reconstruction problem.

\begin{figure}[t]
\centering
\subfloat{\includegraphics[trim=3cm 3.5cm 4cm 3.2cm, clip, width=4.4in]{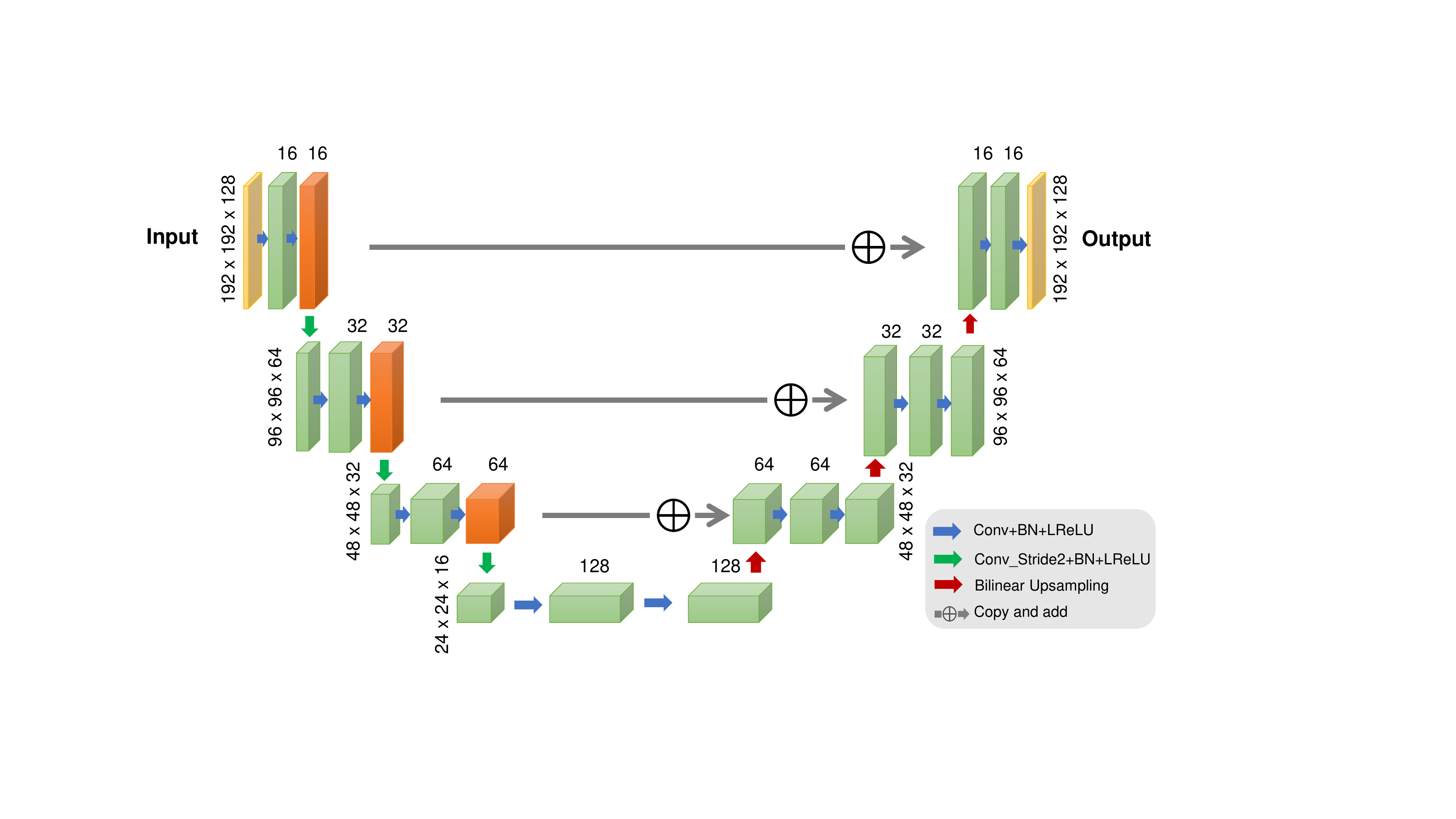}}\vspace{-0.1in}
\caption{\small{The schematic plot of the Network structure used in this work. The spatial input size for each layer is based on the real data experiment described in Section.~\ref{section:brain_pet}.} }
\label{fig:3d_unet}
\end{figure}
\begin{figure}[t]
\centering
\subfloat{\includegraphics[trim=0cm 0cm 0cm 0cm, clip, width=2.5in]{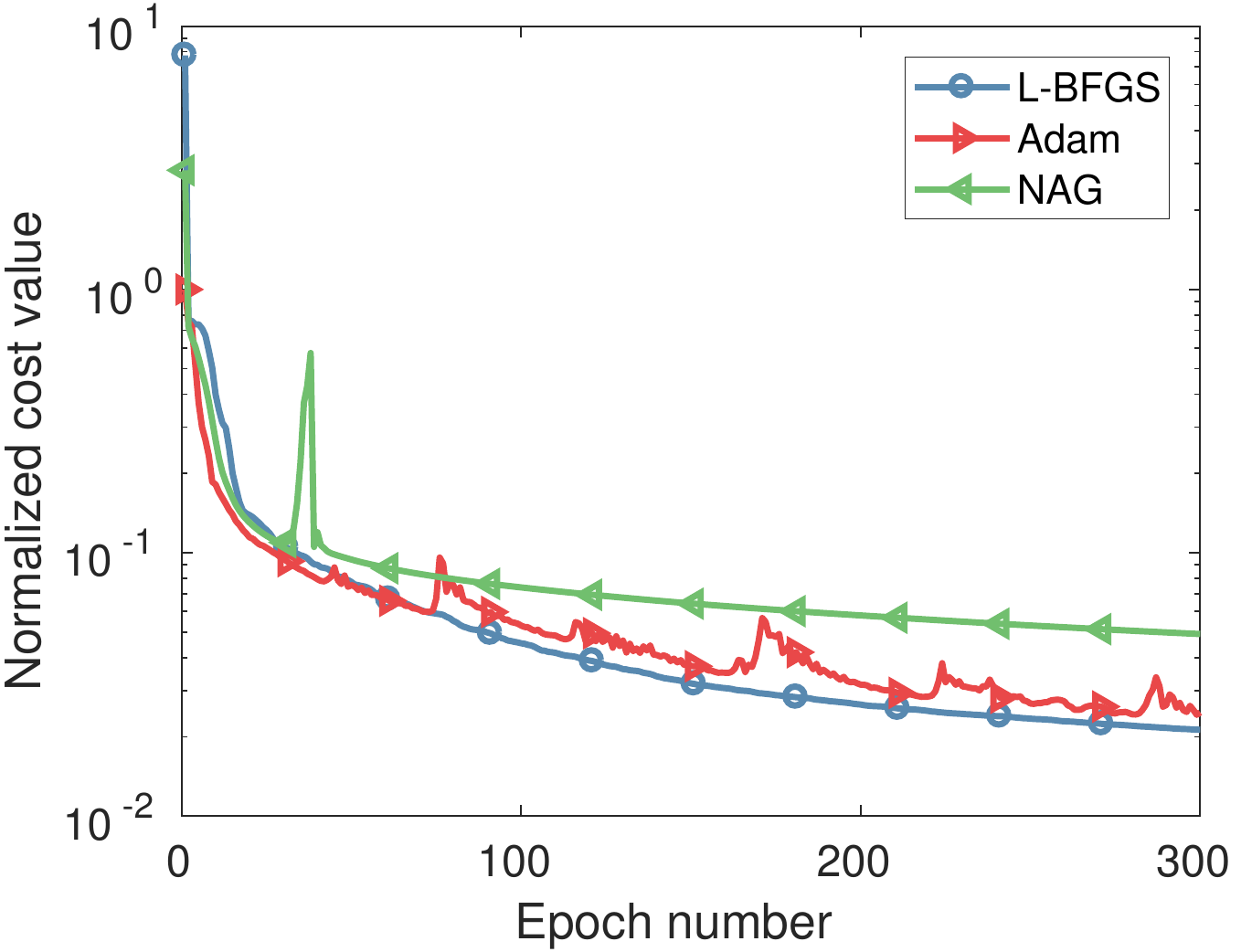}}
\subfloat{\includegraphics[trim=0cm 0cm 0cm 0cm, clip, width=2.5in]{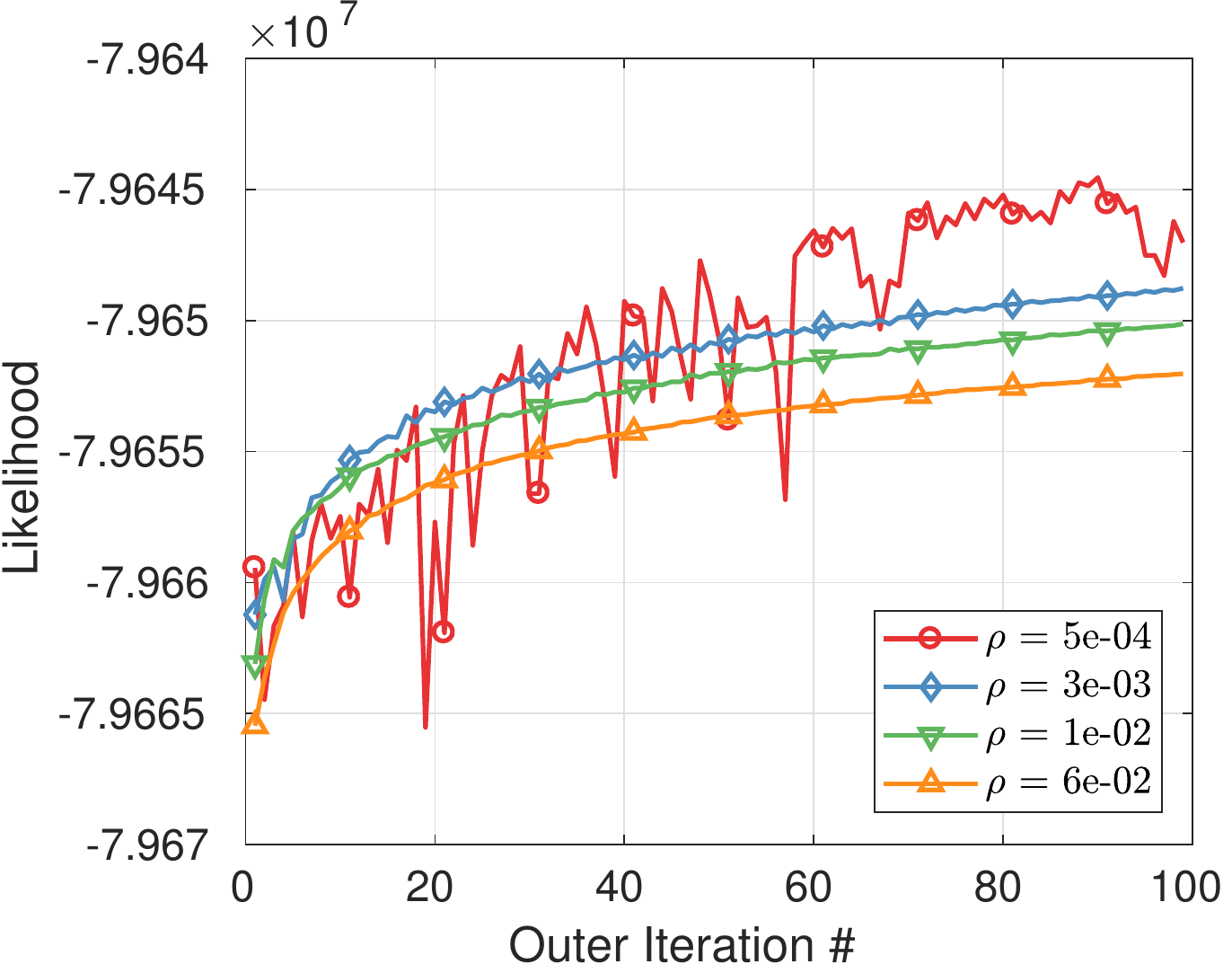}}
\caption{\small{(Left) Comparison of the normalized likelihood for the Adam, Nesterov’s accelerated gradient (NAG) and L-BFGS algorithms. (Right) The effect of penalty parameter $\rho$ on the likelihood $L(\bb{y} | f(\bb{\hat{\theta}} | \bb{\alpha}))$}. }
\label{fig:effect_like_rho}
\end{figure}
\subsubsection{Network structure and implementation details}\label{section:implement_details}
The network structure employed in this work is based on the modified 3D U-net \citep{cciccek20163d} as shown in Fig.~\ref{fig:3d_unet}. In our implementation, there are several modifications compared to the original 3D U-net: (1) using convolutional layer with stride 2 to down-sample the image instead of using max pooling, to construct a fully convolutional network; (2) directly adding the left side feature to the right side instead of concatenating, to reduce the number of training parameters, thus the risk of overfitting; (3) using the bi-linear interpolation instead of the deconvolution upsampling, to reduce the checkboard artifact; (4) using leaky ReLU instead of ReLU. We compared the behaviors of different network training algorithms for subproblem (\ref{eq:sub1}). The Adam,  NAG, and L-BFGS algorithms were compared after running 300 epochs using the real brain data mentioned in Section.~\ref{section:brain_pet}. When comparing different algorithms, we computed the normalized cost value, which is defined as $L_n = (\phi_{\text{Adam}}^{\text{ref}} - \phi^{n})/(\phi_{\text{Adam}}^{\text{ref}} - \phi_{\text{Adam}}^{1})$,where $\phi_{\text{Adam}}^{\text{ref}}$ and $\phi_{\text{Adam}}^{1}$ is the cost value after running Adam for 700 iterations and 1 iteration, respectively.  Fig.~\ref{fig:effect_like_rho}(a) plots the normalized cost value curves for different algorithms. The L-BFGS algorithm is monotonic decreasing while the Adam algorithm is not due to the adaptive learning rate implemented. The NAG algorithm is slower than the other two algorithms. Due to the monotonic property, the network output using the L-BFGS algorithm is more stable and less influenced by the image noise when running multiple realizations. This is the reason we chose L-BFGS algorithm to solve subproblem (\ref{eq:sub1}). All the neural network related training was implemented using TensorFlow 1.4. As the penalty parameter $\rho$ used in the ADMM framework has a large impact on the convergence speed, we examined the log-likelihood $L(\bb{y} | f(\bb{\hat{\theta}} | \bb{\alpha}))$  to determine the penalty parameter used in practice. As an example, Fig.~\ref{fig:effect_like_rho} (b) shows the log-likelihood curves using different penalty parameters for the real brain data mentioned in Section.~\ref{section:brain_pet}. Considering the convergence speed and stability of the likelihood, $\rho$ = 3e-3 was chosen. 

\section{Experiment}
\subsection{Brain PET image reconstruction: simulation study}\label{section:brain_pet_simu}
\begin{figure}[t]
\centering
\subfloat{\includegraphics[trim=2.4cm 0.5cm 2.3cm 0.3cm, clip, width=4.5in]{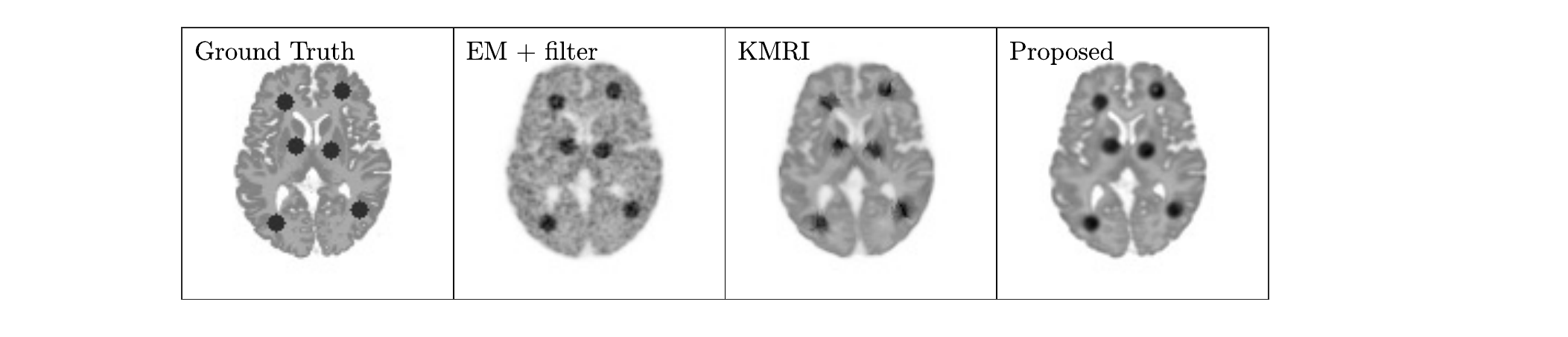}}\vspace{-0.1in}\\
\subfloat{\includegraphics[trim=2.4cm 0.4cm 2.3cm 0.2cm, clip, width=4.5in]{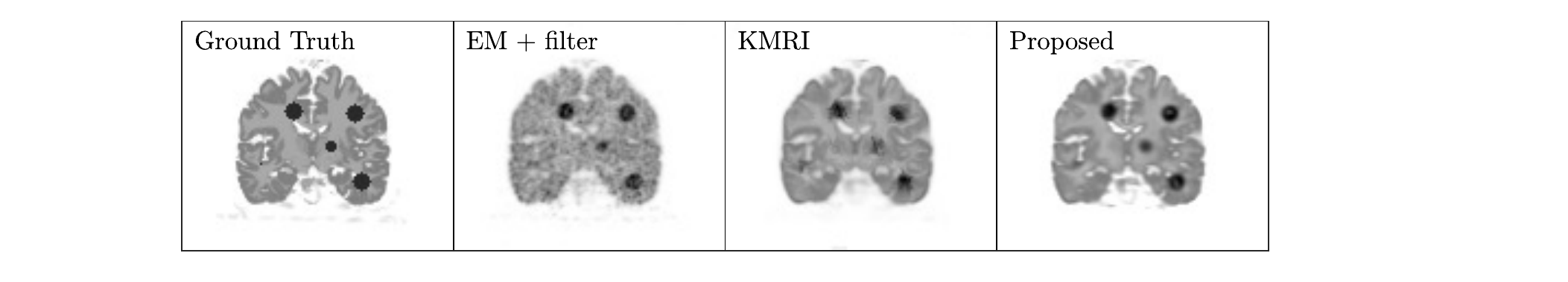}}\vspace{-0.1in}\\
\subfloat{\includegraphics[trim=2.4cm 0.4cm 2.3cm 0.2cm, clip, width=4.5in]{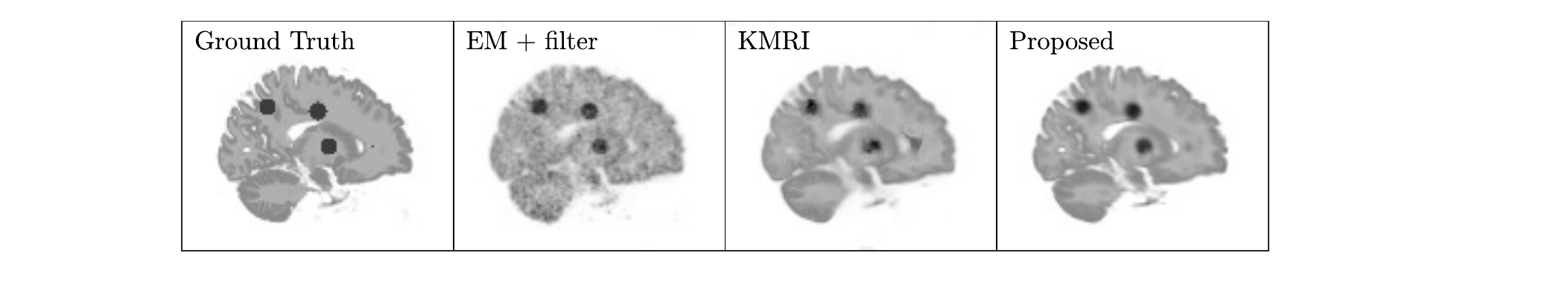}}\\
\caption{Three orthogonal slices of reconstructed images using different methods for the simulation brain dataset.}
\label{fig:simu_img_appear}
\end{figure}

A 3D brain phantom from BrainWeb \cite{aubert2006twenty} was used in the simulation. Corresponding T1 weighted MR image was used as the prior image. The voxel size is 2$\times$2$\times$2 $\mbox{mm}^3$ and the phantom image size is 128$\times$128$\times$105. 
To simulate mismatches between the MR and PET images, twelve hot spheres of diameter 16 mm were inserted into the PET image as tumor regions, which are not visible in the MR image. In this experiment, the last 5 min frame of a one-hour FDG scan was used as the ground-truth image. The computer simulation modeled the geometry of a Siemens mCT scanner. Noise-free sinogram data were generated by forward-projecting the ground-truth image using the system matrix and the attenuation map. Poisson noise was then introduced to the noise-free data by setting the total count level to be equivalent to last 5 min scan with 5 mCi injection.  Gaussian post-filtering method (denoted as EM+filter) and the kernel method (denoted as KMRI) \citep{guobao15} were employed as comparison methods. Fig.~\ref{fig:simu_img_appear} shows three orthogonal views of the reconstructed images using different methods. The kernel method and the proposed method both reveal more cortex structures compared to the EM-plus-filter method due to the boundary information provided by the MR priors. Compared to the kernel method, the proposed method can recover even more details of the cortices and the white matter regions are cleaner. Besides, the tumor uptake using the proposed method is higher and the tumor shape is closer to the ground truth. This means that even when there are mismatches between the PET and MR images, the proposed method can still recover the true PET intensities and shapes. Fig.~\ref{fig:real_crc_std}(a,b) shows the contrast recovery coefficient(CRC). vs standard deviation (STD) curves for different methods. For both the gray matter region and the tumor region, the proposed method out-performs other methods. 

\subsection{Brain PET image reconstruction: clinical data study}\label{section:brain_pet}
\begin{figure}[t]
\centering
\subfloat{\includegraphics[trim=0cm 1.5cm 2.8cm -0.5cm, clip, width=1.27in]{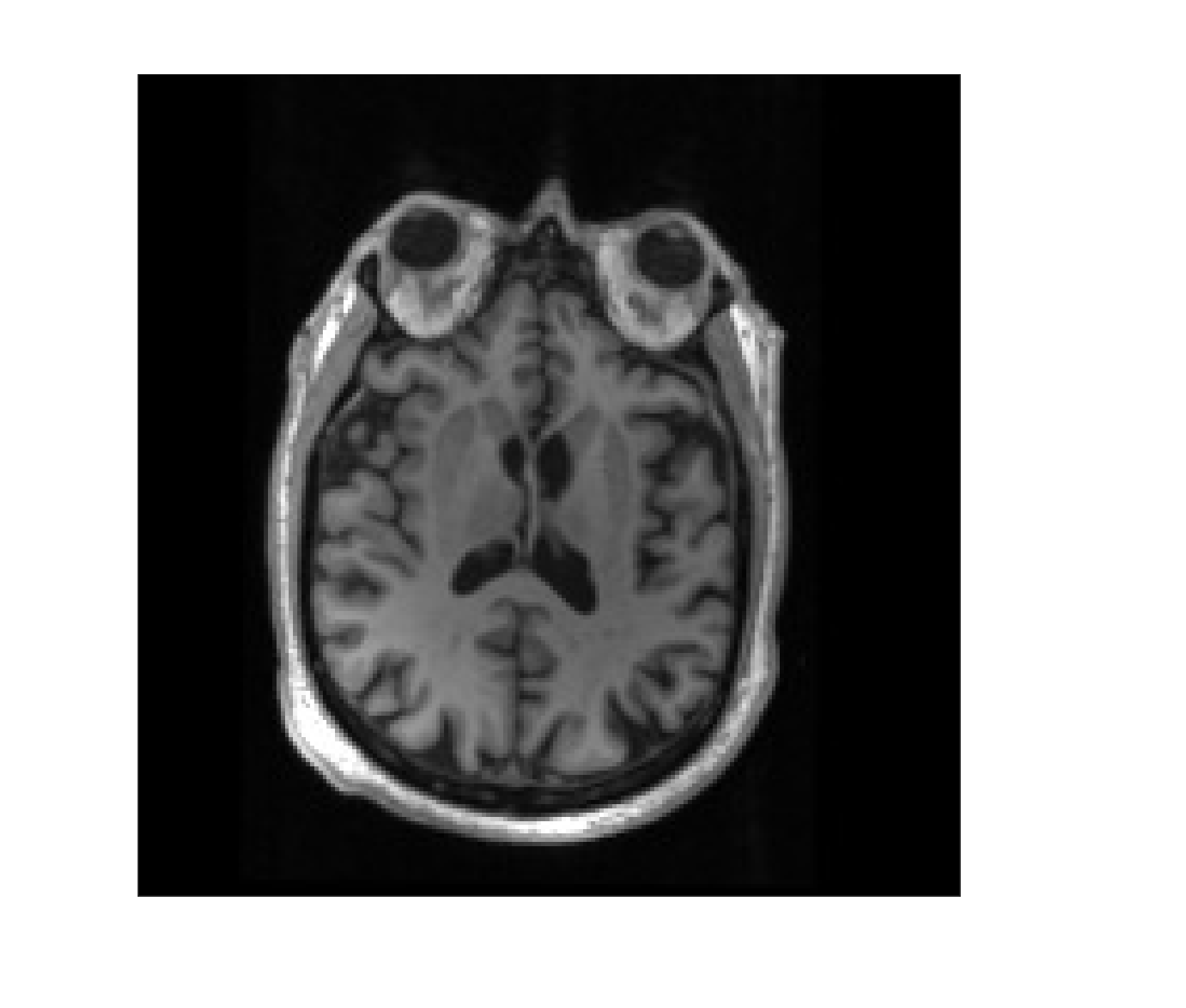}}
\subfloat{\includegraphics[trim=2.8cm 0.75cm 2cm 0.5cm, clip, width=3.6in]{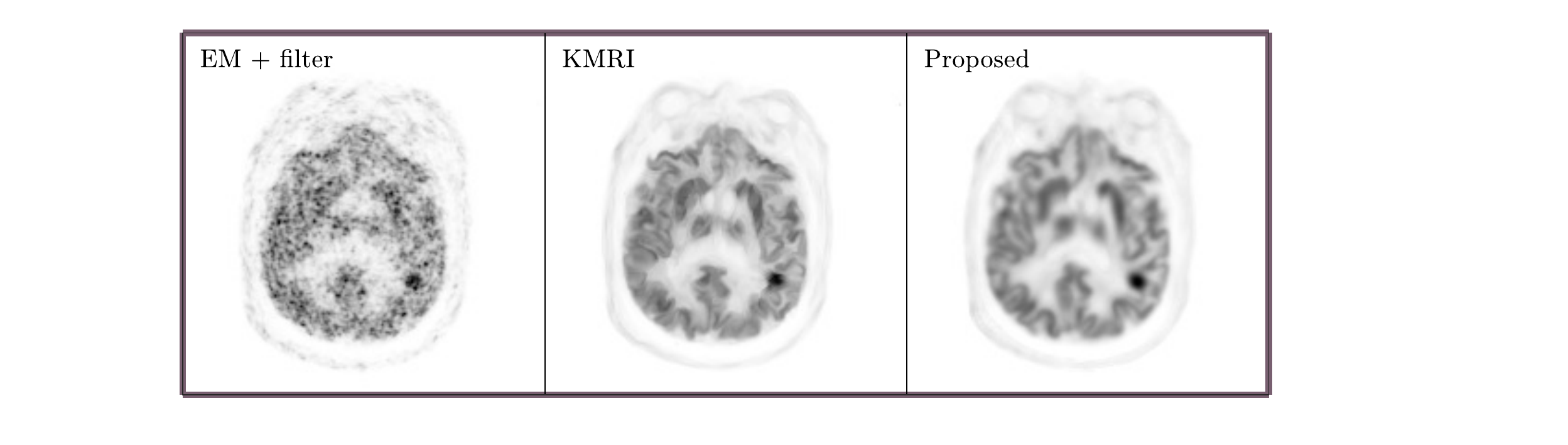}}\vspace{-0.1in}\\
\subfloat{\includegraphics[trim=0cm 1.4cm 2.8cm 0.3cm, clip, width=1.27in]{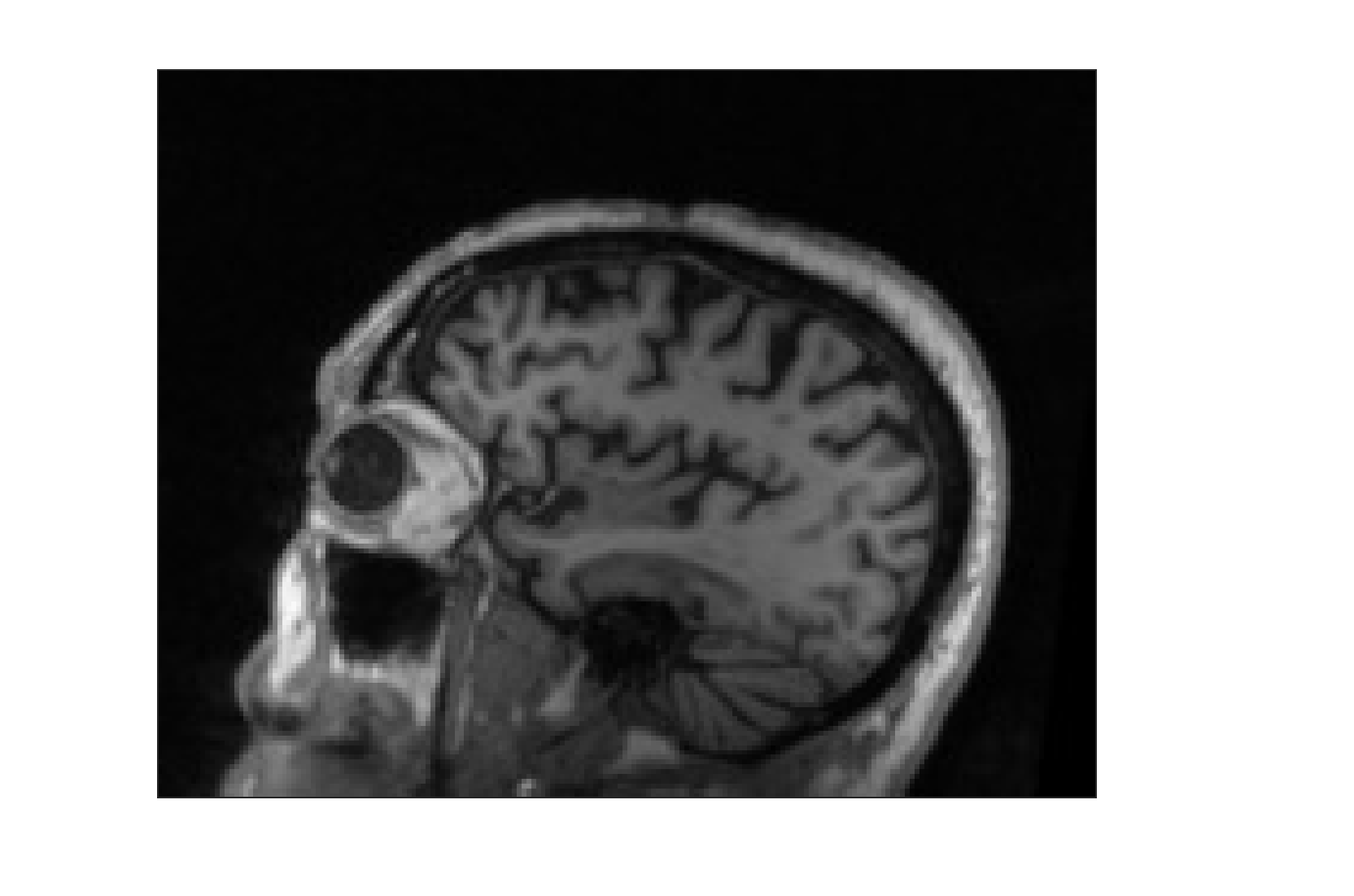}}
\subfloat{\includegraphics[trim=2.8cm 0.6cm 2cm 0.35cm, clip, width=3.6in]{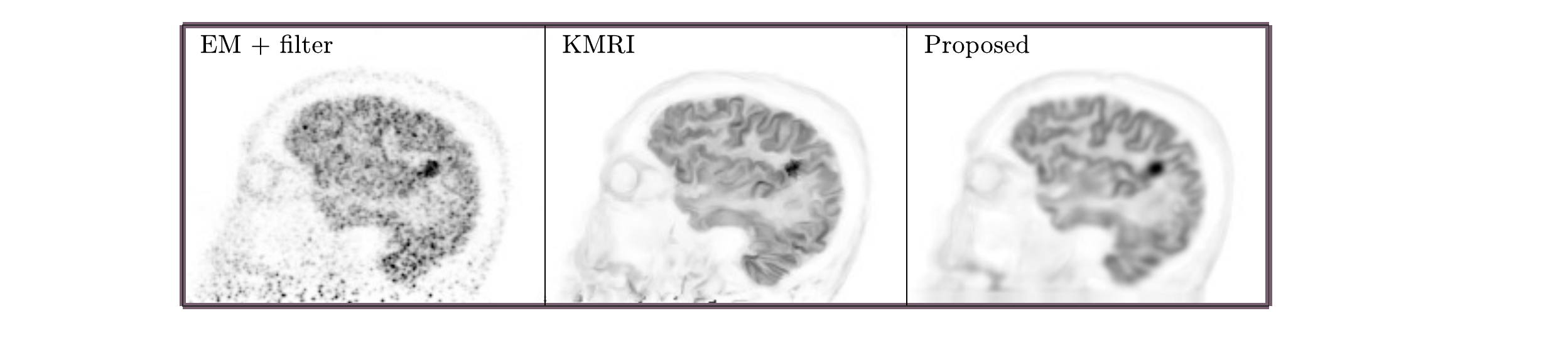}}
\vspace{-0.1in}\\
\subfloat{\includegraphics[trim=0cm 1.4cm 2.8cm 0.3cm, clip, width=1.27in]{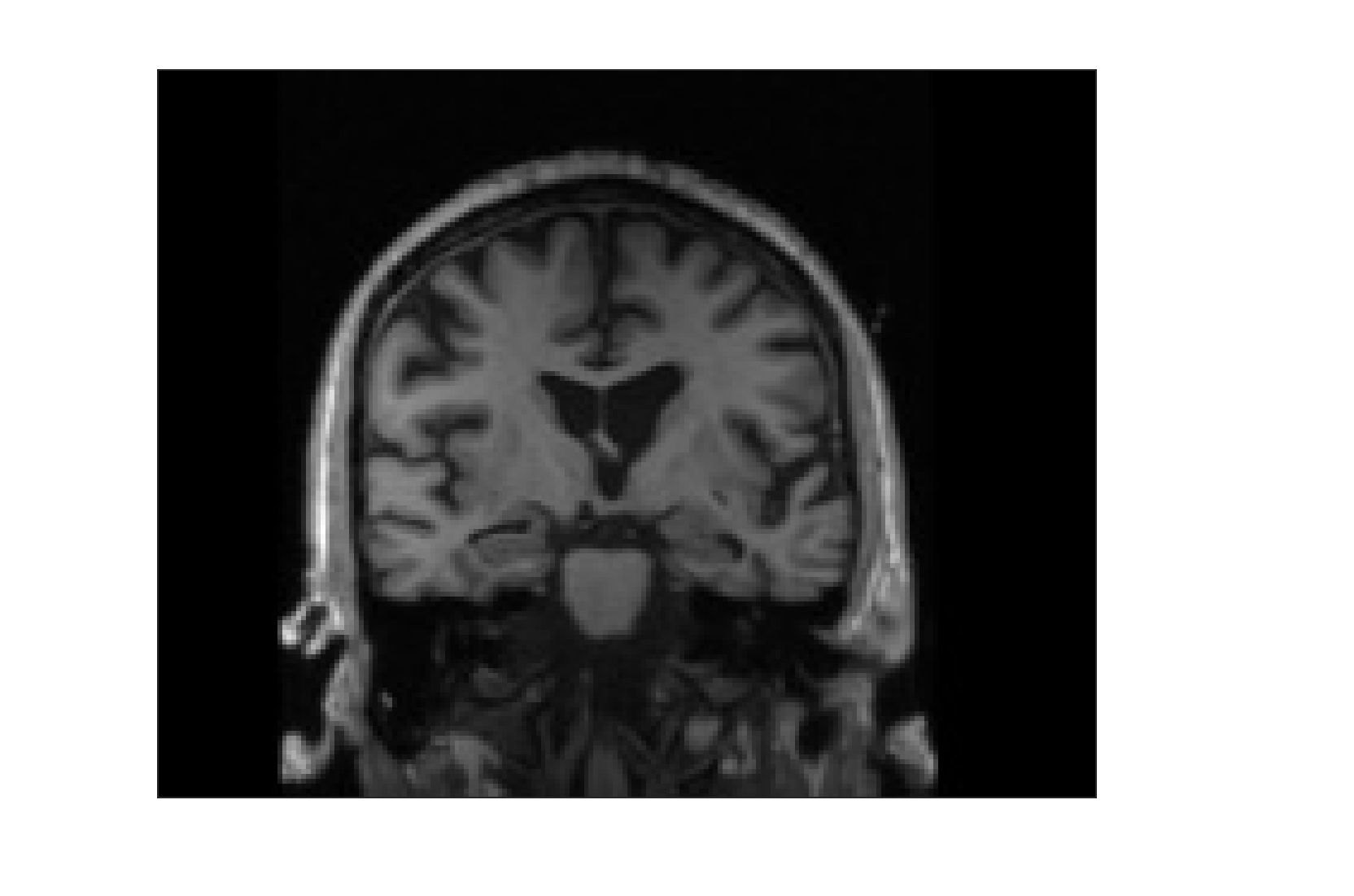}}
\subfloat{\includegraphics[trim=2.8cm 0.6cm 2cm 0.35cm, clip, width=3.6in]{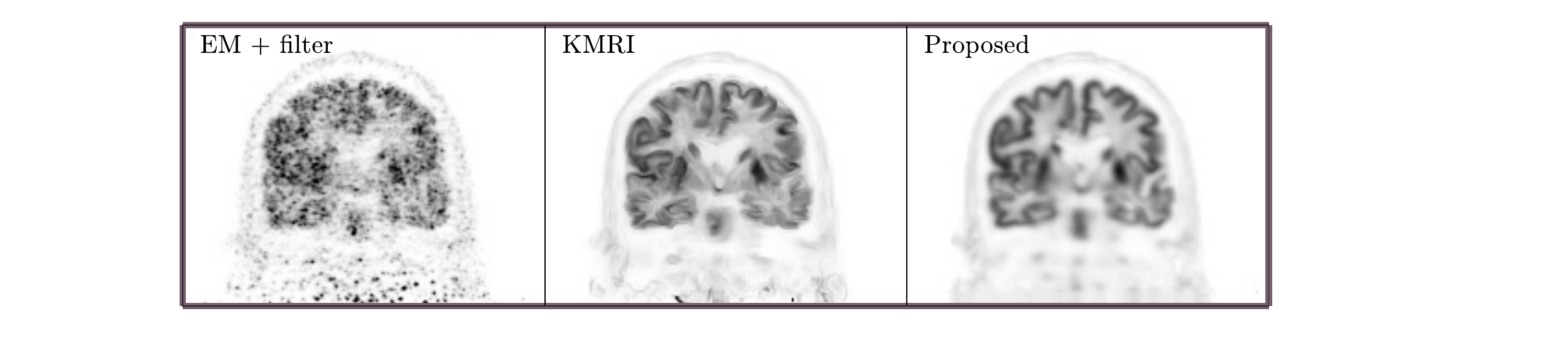}} \\
\caption{\small{The three rows show three orthogonal slices  for the clinical brain dataset using different method. The first column is the corresponding MR prior image.}}
\label{fig:real_img_appear}
\end{figure}
\begin{figure}[h]
\centering
\subfloat{\includegraphics[trim=0cm 0cm 0cm 0cm, clip, width=1.75in]{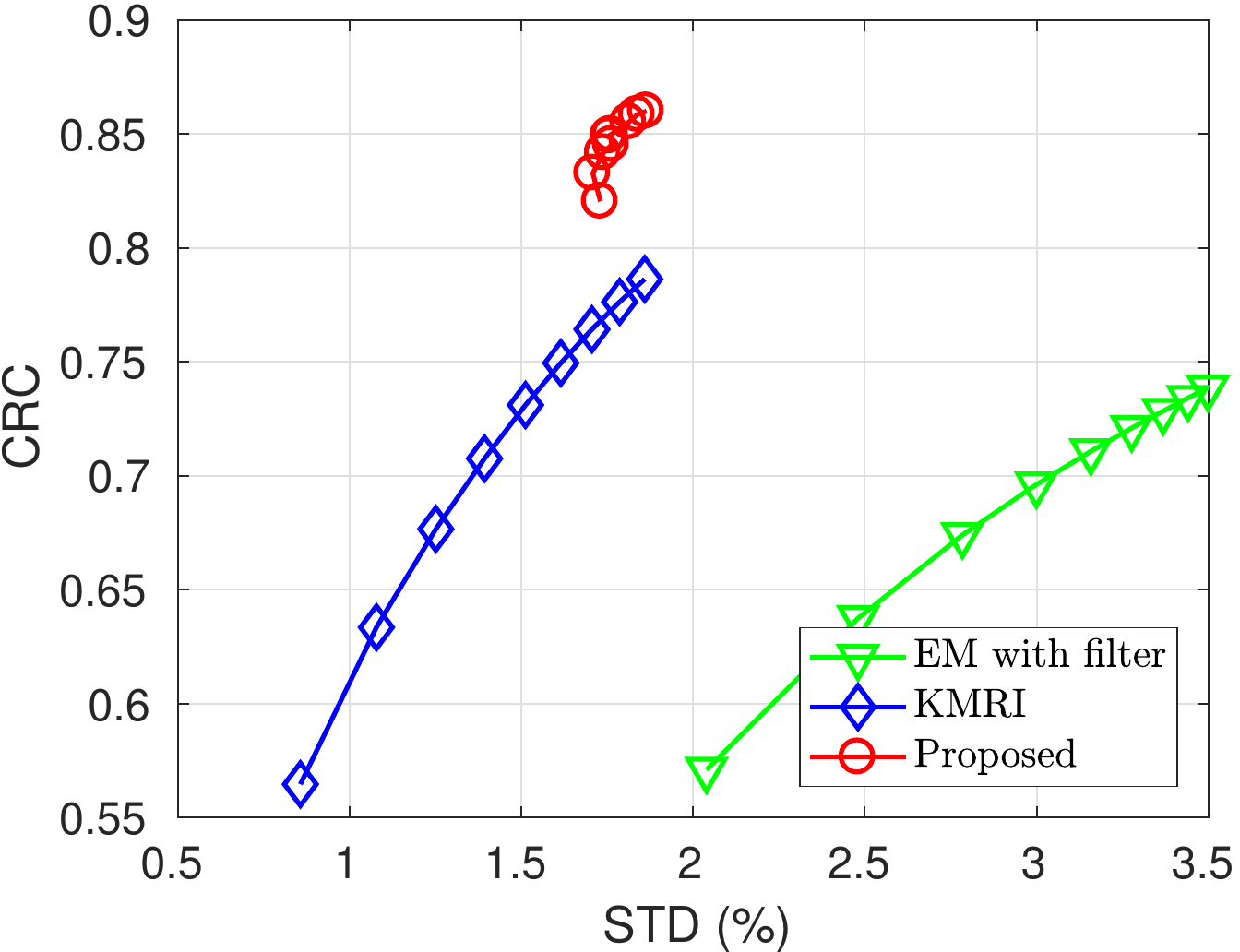}}
\subfloat{\includegraphics[trim=0cm 0cm 0cm 0cm, clip, width=1.75in]{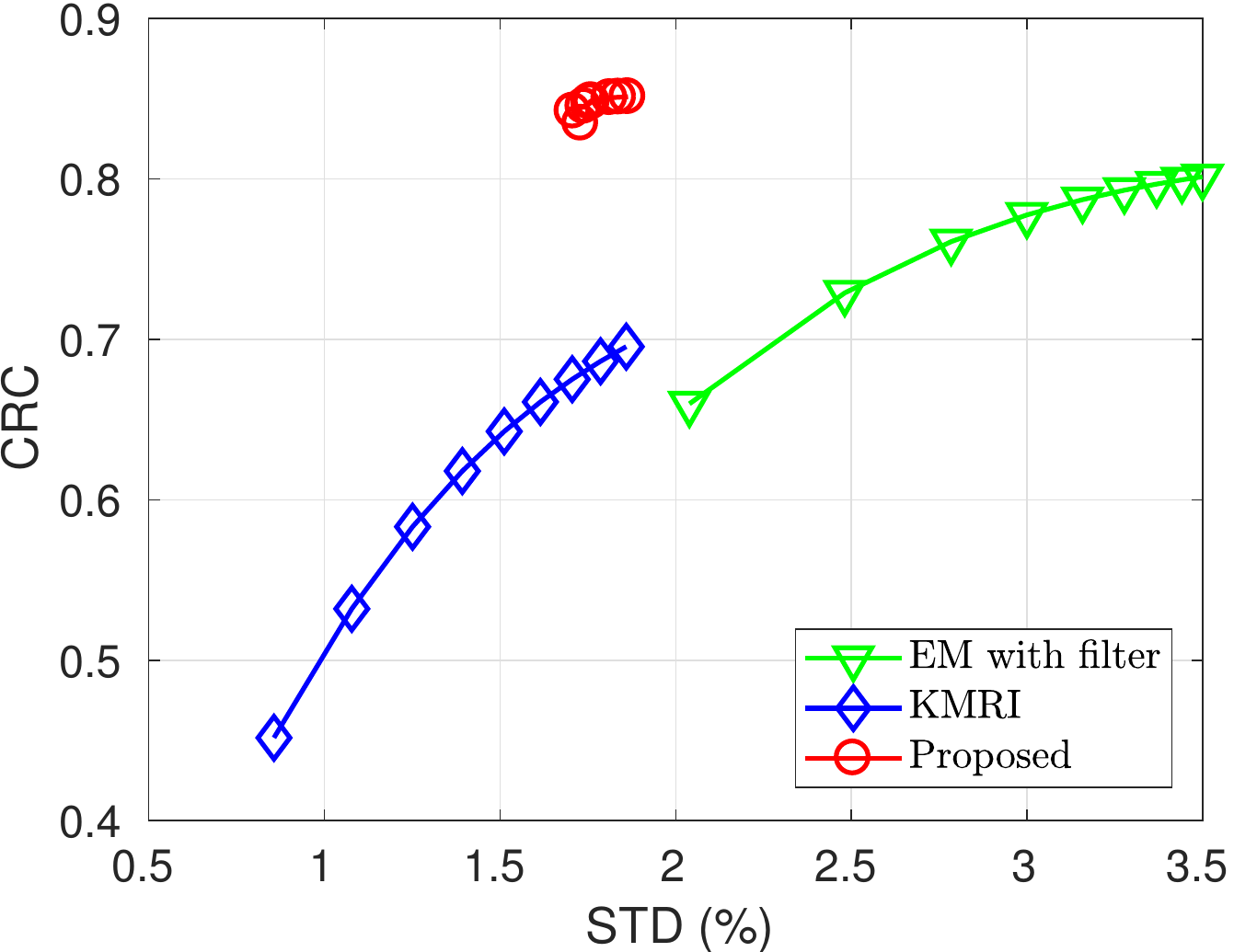}}
\subfloat{\includegraphics[trim=0cm 0cm 0cm 0cm, clip, width=1.75in]{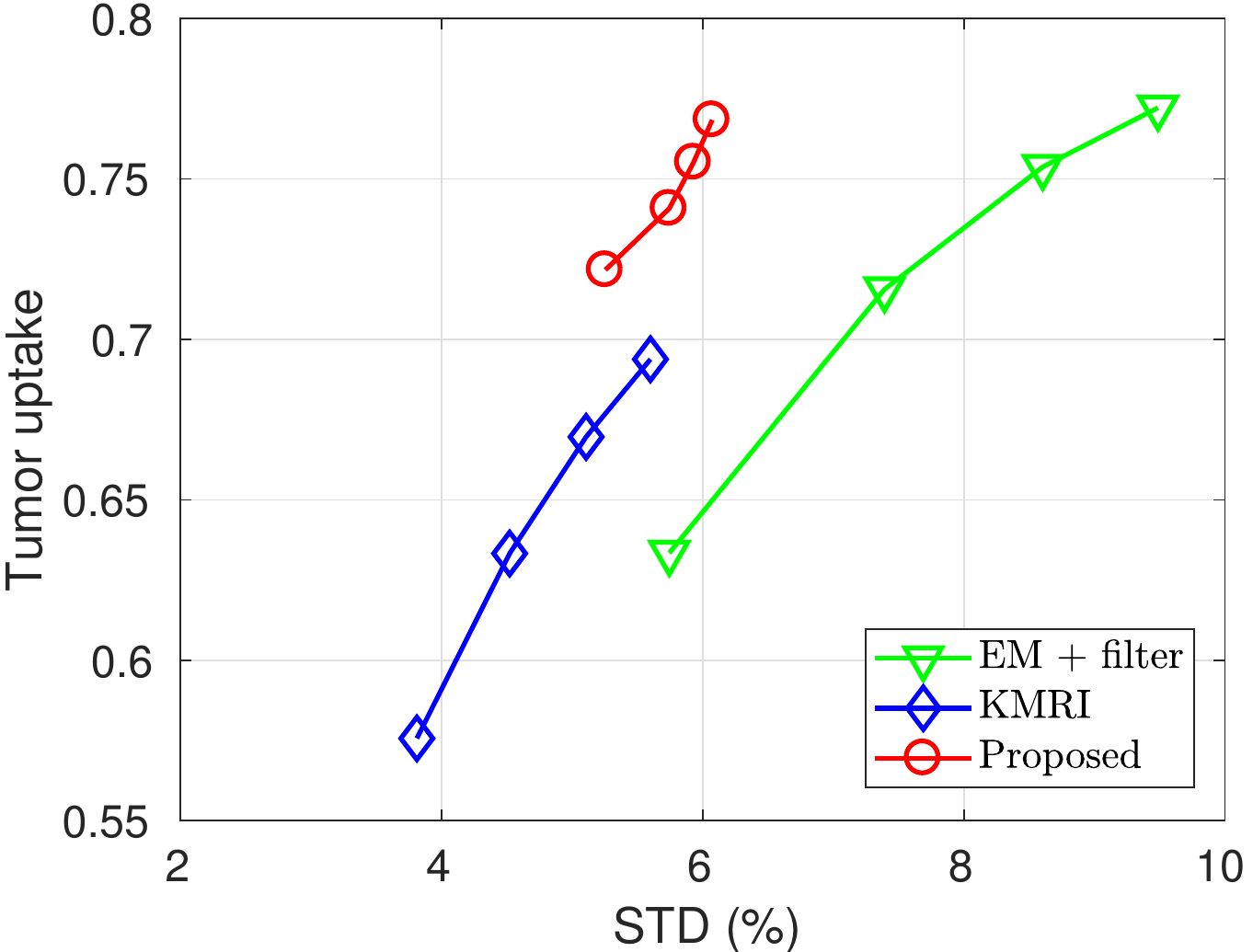}}
\caption{\small{(Left) The CRC-STD curve for gray matter region in the brain simulation study; (Middle) The CRC-STD curve for tumor region in the brain simulation study. (Right) The CR-STD curve for tumor region in the clinical brain dataset. Markers are generated for every twenty iterations.}}
\label{fig:real_crc_std}
\end{figure}
A 70-minutes dynamic PET scan of a human subject acquired on a Siemens Brain MR-PET scanner after 5 mCi FDG injection was employed in the real data evaluation. The data were reconstructed with an image array of 256$\times$256$\times$153 and a voxel size of 1.25 $\times$ 1.25 $\times$ 1.25 $\mbox{mm}^3$. A simultaneous acquired {T1-weighted} MR image has the same image array and voxel size as the PET images. Correction factors for attenuation, randoms, scatters were estimated using the standard software provided by the manufacturer and included during reconstruction. The motion correction was performed in the LOR space based on the simultaneously acquired MR navigator signal\citep{catana2011mri}. To generate multiple realizations for quantitative analysis, the last 40 minutes PET data were binned together and resampled with a 1/8 ratio to obtain 20 i.i.d. datasets that mimic 5-minutes frames. As the ground truth of the regional uptake is unknown, a hot sphere with diameter 12.5 mm, mimicking a tumor, was added to the PET data (invisible in the MRI image). For tumor quantification, images with and without the inserted tumor were reconstructed and the difference was taken to obtain the tumor only image and compared with the ground truth. The tumor contrast recovery (CR) was calculated as ${\text{CR} = 1/R\sum\nolimits_{r = 1}^{R}\bar{l}_{r} / {l}_{\text{true}}}$, where $\bar{l}_{r}$ is the mean tumor uptake inside the tumor ROI, ${l}_{\text{true}}$ is the ground truth of the tumor uptake, and $R$ is the number of the realizations. For the background, 11 circular ROIs with a diameter of 12.5 mm were drawn in the white matter to calculate the standard deviation. Fig.~\ref{fig:real_img_appear} shows the reconstructed images and the corresponding MR prior images of the real brain dataset using different methods. For the methods with MR information included, more cortex details are recovered and the image noise in the white matter is much reduced. The cortex shape in the proposed method is clearer than the kernel method. For the tumor region which is unobserved in the MR image, the uptake is higher in the proposed method compared with the kernel method. Fig.~\ref{fig:real_crc_std}(c) shows the CR-STD curves for different methods. Clearly the proposed method has the best CR-STD trade-off compared with other methods.

\subsection{Clinical PET image denoising}
\begin{figure}[t]
\centering
\subfloat{\includegraphics[trim=0cm 0cm 0cm 0cm, clip, width=5in]{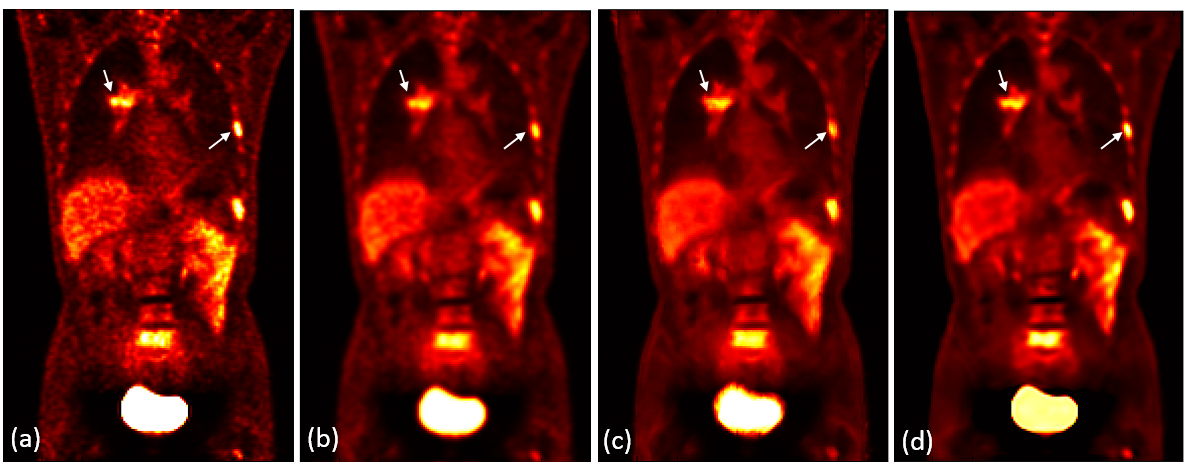}}
\caption{\small{Coronal view of (a) original noisy PET image; (b) post-filtered image using a Gaussian filter with FWHM = 1 pixel; (c) post-filtered image using NLM with CT images and window size set to $5 \times 5 \times5$; (d) post-filtered image using the proposed method trained with 700 epochs. Lesions are pointed out using arrows.}}
\label{fig:image}
\end{figure}

The patient images were acquired with a Siemens Biograph mCT PET/CT system. Patients were injected with 111 MBq (3 mCi) of 68Ga-PRGD2. PET image at 60 min post injection was acquired with 5 bed positions.  For the proposed method, the corresponding CT image is employed as the network input. The Gaussian denoising, and the non-local mean (NLM) filtering guided by corresponding CT images \cite{chan2014postreconstruction} were employed as the reference methods. To evaluate the performance of different methods quantitatively, the contrast-to-noise ratio (CNR) regarding the lesion and muscle regions was used as the figure of merit, which is defined as $\text{CNR} = ({{m}_{\text{lesion}}-{m}_{\text{mucsle}}})/{{{\sigma}}_{\text{muscle}}}$, where ${m}_{\text{lesion}}$ and ${m}_{\text{muscle}}$ denote the mean intensity inside the lesion ROI and muscle ROIs, respectively, and ${\sigma}_{\text{muscle}}$ is the pixel-to-pixel standard deviation inside the chosen muscle ROIs. Fig.~\ref{fig:image} shows the coronal view of one patient data processed using different methods given the optimal parameter for each method. It can be observed that the proposed method has the best visual appearance, which smooths out the noise but also keeps the detailed organ and tumor structures. Table \ref{table_2} shows the CNRs for all the patient datasets using different methods. The proposed method has the best performance for all the patient datasets. In this example, clinical whole-body PET denoising were employed to show the superior performance of the proposed framework quantitatively. When applied to brain imaging denoising applications, the proposed framework can also perform well. Fig.~\ref{fig:effect_of_conditional} is a qualitative example. More quantitative evaluations are needed for brain imaging restoration applications.  

\begin{table}[h]
\renewcommand{\arraystretch}{1}
\caption{\small{CNRs of ten patient datasets using Gaussian method, NLM method, and the proposed method.}}
\label{table_2}
\centering
\begin{tabular}{c|c|c|c|c|c|c|c|c|c|c}
	\hline\hline
	Patient index & 1 & 2 & 3 & 4 & 5 & 6 & 7 & 8 & 9 & 10 \\
	\hline
	Gaussian & $41.11$ &  $3.61$ &  $7.11$&  $4.51$& $3.61$ & $9.17$& $17.40$&  $2.13$ &  $0.83$& 	 $5.07$ \\
	\hline
	NLM & $45.18$& $3.64$ & $7.38$  &  $4.54$& $4.42$ & $9.19$ &  $20.77$ &  $2.39$& $0.88$& $4.43$ \\
	\hline
	Proposed & $\bb{49.29}$& $\bb{5.22}$ &$\bb{7.58}$ & $\bb{5.29}$& $\bb{4.59}$ &$\bb{9.87}$ &  $\bb{22.60}$& $\bb{2.76}$& $\bb{1.28}$& $\bb{6.38}$  \\
\hline\hline
\end{tabular}
\end{table}

\section{Discussion and Conclusion}
In this work, we propose a personalized learning framework for inverse problems in medical imaging using deep neural network. Prior training pairs are not needed in this process, but only the patient's own prior images. An ADMM framework is proposed to  decouple the whole optimization into a penalized inverse problem and a network training problem. Brain PET image reconstruction and denoising were employed as examples to demonstrate the effectiveness of this proposed framework. Quantitative results show that the proposed personalized representation framework performs better than other widely adopted methods. One advantage of decoupling the whole optimization is that all current algorithms for inverse problem and network training can be employed, which makes the proposed framework more adoptable. One weakness of the proposed framework is its lack of theoretical convergence guarantee though monotonic convergence is observed in Fig.~\ref{fig:effect_like_rho}(b). U-net structure is employed in the framework due to its strong representation ability, so that the number of trainable network parameters is less than the original inverse problem. Finding a network structure with better representation ability will improve the performance of the proposed framework. Apart from PET imaging, this proposed framework can also be extended to other inverse problems in medical imaging where personalized priors are available, such as dynamic brain imaging, longitudinal imaging, and multi-parametric imaging scenarios. The superior performance of the proposed method is essentially due to a better representation of image. Future work will focus on theoretical convergence analysis and more thorough clinical evaluations.

\end{document}